\documentclass[conference]{IEEEtran}
\IEEEoverridecommandlockouts
\usepackage{cite}
\usepackage{amsmath,amssymb,amsfonts}
\usepackage{algpseudocode}
\usepackage{booktabs}
\usepackage{url}
\usepackage{graphicx}
\usepackage{textcomp}
\usepackage{xcolor}
\usepackage{float}
\usepackage{listings}
\usepackage{caption}
\usepackage{algorithm}
\def\BibTeX{{\rm B\kern-.05em{\sc i\kern-.025em b}\kern-.08em
    T\kern-.1667em\lower.7ex\hbox{E}\kern-.125emX}}
\begin{document}

\title{Consistent Relexicalization of Clinical Documents using Graph-Based Approach\\
}

\author{
\IEEEauthorblockN{
Dipankar Das,
Atri Mandal,
Sandeep Singh,
Tushar Shandhilya
}
\IEEEauthorblockA{
\textit{Oracle Health AI} \\
Bengaluru, India \\
\{dipankar.das, atri.mandal, sandy.singh, tshandhilya.shandhilya\}@oracle.com
}
}


\maketitle

\begin{abstract}
Relexicalization is a pivotal technique in clinical NLP, as it facilitates robust masking of sensitive information while synthesizing datasets that retain high-fidelity, real-world characteristics. However, preserving structural integrity, relational coherence, and temporal consistency during transformation remains a significant challenge. Existing approaches frequently rely on independent entity replacement, which results in clinical inconsistencies across longitudinal records. This reduces the value of such relexicalized datasets for downstream scientific analysis. To address these limitations, we introduce G-RELIC (\textbf{G}raph Based Contextual \textbf{REL}exicalization with \textbf{I}mproved \textbf{C}onsistency) which combines the power of LLMs with graphs. \\
G-RELIC implements a graph-based mapping mechanism which optimizes for one-to-one correspondence between original and surrogate entities. It also introduces a deterministic temporal repositioning algorithm to preserve temporal consistency. \\ 
Empirical evaluations on diverse, real-world clinical datasets validate that G-RELIC significantly outperforms state-of-the-art baselines. G-RELIC yields a 30.4 percentage point improvement in relational integrity (62.1\% to 92.5\%) and 45.9 percentage point improvement in temporal coherence (46\% to 91.9\%) without compromising on the recognized privacy benchmarks for clinical datasets. This maximizes the analytical utility of relexicalized datasets while minimizing re-identification risk. 
\end{abstract}


\section{Introduction}
\label{sec:intro}
The digital transformation of healthcare has generated vast repositories of Electronic Health Records (EHRs), offering unprecedented opportunities for downstream clinical research, such as disease phenotyping and predictive modeling. However, the sharing of these datasets is strictly governed by privacy regulations (e.g., HIPAA, GDPR), necessitating robust de-identification. While simple redaction or masking of entities ensures privacy, it removes valuable clinical context, rendering the de-identified datasets unusable for scientific analysis. \\
Relexicalization (relex) addresses this problem by replacing protected health information (PHI)/personally identifiable information (PII) with realistic surrogates and is critical for creating realistic synthetic datasets for researchers without risking patient privacy. However, most existing methods of relexicalization treat clinical documents as isolated snapshots rather than parts of a continuous narrative. This leads to severe limitations when processing longitudinal records, which comprise a series of clinical encounters for a single patient over months or years. In these records, the preservation of interconnections is essential for any meaningful scientific analysis.
\subsection{The Challenge of Consistency}
\label{sec:challenges}
In longitudinal relexicalization, the primary hurdle is maintaining consistency across multiple documents. Without a mechanism to anchor entities and timestamps, independent relexicalization leads to two major types of structural degradation: \\
\textbf{Relational Integrity:} 
For non-temporal entities (e.g., clinicians, facilities, and family members), inconsistent mapping results in: \par
\textit{Identity Fragmentation:} Occurs when a single real-world entity (e.g., "Dr. Alice Sun") is replaced by different surrogates (e.g., "Dr. Bob" in Document A and "Dr. Charlie" in Document B). This breaks the link between a patient and their primary care provider across encounters.\par
\textit{Identity Merging:} Occurs when two distinct entities are mapped to the same surrogate, erroneously suggesting a relationship that does not exist.\\
\textbf{Temporal Inconsistency:} Traditional de-identification methods, which often treat timestamps in isolation and rely on independent random date shifting, suffer from critical limitations resulting in unrealistic timelines. For example, events like a patient's discharge may be shifted to precede the admission date, or specific dates may be altered while relative markers, like mentioned weekdays, remain static. Other inconsistencies may include loss of precise intervals between clinical events (e.g., duration between drug administration and subsequent lab results) or a loss of medical plausibility due to age-disease correlations being severed (such as diagnosis of dementia in early childhood). \\ 
Such inconsistencies may render the transformed data useless for pharmacological and longitudinal research.
\subsection{Main Contributions}
To address these gaps, we propose G-RELIC, a graph-grounded framework designed to generate high-fidelity, research-ready clinical data. The proposed system makes the following key contributions: \\
\textbf{Graph-Driven Framework for High-Fidelity Data Generation:} G-RELIC leverages the power of LLMs combined with a \textbf{Clinical Knowledge Graph (CKG)} to relexicalize clinical records. This framework captures the intricate relationships between patients, providers, and medical events, ensuring that the generated output retains the structural complexity of real-world clinical data required for clinical research.\\
 \textbf{Multi-tier Entity Resolution for Better Consistency:} We implement a multi-tier entity resolution technique within the CKG which uses a combination of deterministic and probabilistic search techniques to ensure consistent entity normalization across the entire dataset. By preventing identity fragmentation and accidental merging, our method improves relational consistency by 30 percentage points over baseline methods.\\
\textbf{Additional Reliability through Temporal Context Preservation:} We develop a hybrid temporal validation framework to ensure the temporal context of the generated data is grounded in medical plausibility and all time-based shifts are deterministic and sequence-aware, thereby overcoming longitudinal drift and chronological incoherence. 
\begin{figure*}[h]
    \centering
    \includegraphics[width=1.0\textwidth, height=0.3\textheight, keepaspectratio]{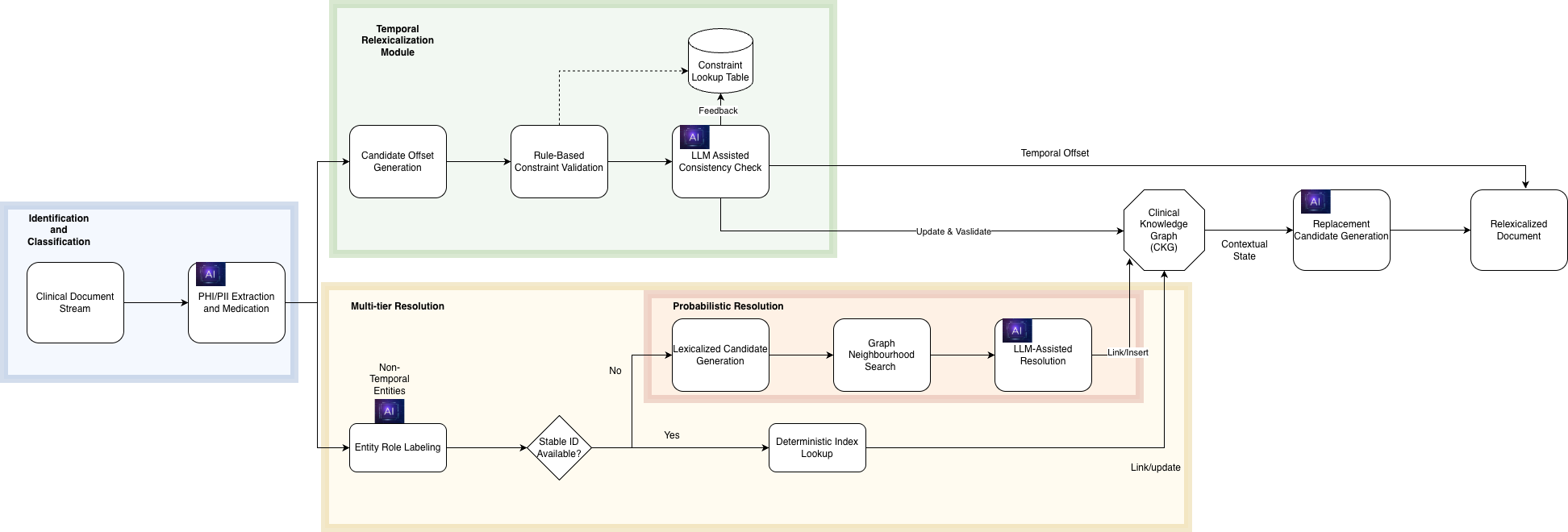}
    \caption{Architecture Diagram for \textbf{G-RELIC}}
    \label{fig:Architecture}
\end{figure*}

\section{Related Work}
\label{sec:relatedwork}
\textbf{Traditional De-identification and Masking}:
Early clinical data privacy efforts focused primarily on redaction (complete removal) and masking (replacing with generic placeholders) \cite{Meystre01}, \cite{Neamatullahetal2008}. These methods are largely built to satisfy the HIPAA Privacy Rule’s Safe Harbor method \cite{hhs_ocr_deid_guidance_2012}, which requires the removal of specific categories of Protected Health Information (PHI). Seminal systems like Philter \cite{norgeot2020philter} demonstrated how rule-based redaction can be scaled to millions of documents. These methods are highly secure and provide clear legal compliance but are time consuming \cite{Negashetal2023}.\\ 
\textbf{Rule-Based Relexicalization and Pseudonymization}:
To bridge the utility gap, researchers moved toward relexicalization, that is, substituting PHI with realistic surrogates \cite{Meystre01}. Notable early work includes \cite{Sweeney1996ReplacingPII} and \cite{SweeneyKAnonymity}, which used localized pattern matching, MIST (MIT De-identification System) \cite{aberdeen2010} and \cite{lison-etal-2021-anonymisation}, which employed a combination of lexicons and regular expressions for pseudonymization. While these systems provide natural-looking text, they operate on a local level and cannot ensure consistency of replacement across longitudinal patient records. \\ 
\textbf{Relational consistency in De-identification}:
Maintaining relational and temporal relationships in Knowledge Graphs  has been well-studied in literature but its use in the task of de-identification is practically non-existent. Knowledge graphs (KGs) enable entity linking, relation extraction, and medical entity normalization \cite{Lebleyetal2018}. \cite{ijcai2019p689} investigates the problem of entity resolution in graphs using appropriate similarity measures for different attributes. 
ReLink \cite{9458710} introduces the notion of complete linkage over attributes and hybrid feature spaces using pre-trained models such as BERT. 
The work in ReFinED \cite{ayoola-etal-2022-refined} uses an entity linking approach which uses fine-grained entity types and entity descriptions. However these methods may not capture the full contextual nuances required for accurate disambiguation in complex texts. Additionally, scaling these approaches to handle extensive knowledge bases with millions of entities may present significant computational challenges.
SPEL \cite{shavarani-sarkar-2023-spel} introduces the use of structured prediction for entity linking which classifies each individual input token as an entity and aggregates the token predictions. 
However, residual inconsistencies can still impact entity linking accuracy. Also, the model's reliance on a fixed candidate set may restrict its ability to link to less frequent entities not present in the predefined set. \\
\textbf{Temporal Consistency}:
An early work by \cite{Tangetal2013} introduced a comprehensive system for extraction of temporal relations and expressions from clinical documents while the work by \cite{styler-iv-etal-2014-temporal} introduced THYME guidelines for annotation of temporal relations in clinical text. For relexicalizing temporal entities the standard technique is date shifting (SANT), where a random offset is applied to all dates \cite{Hripcsaketal2016}. However, more recent evaluations have shown that simple shifting often ignores non-date temporal markers like "weeks after" or "last Tuesday," leading to internal contradictions \cite{LIU2017S34}. The work by \cite{evans2023detection} also makes similar observations regarding date-shifting and the necessity for sophisticated relexicalization techniques. \\
\textbf{LLM-Based Approaches}:
The recent emergence of Large Language Models has shifted the paradigm toward zero-shot de-identification. \cite{Wiestetal2025} proposed a Llama-3 based model for clinical relexicalization. Another work - \cite{Kimetal2024} - proposed the use of GPT-4 to generate synthetic clinical context, going beyond simple replacement of entities. Another recent paper, RedactOR \cite{singh-etal-2025-redactor} leverages LLM prompts for both extraction and relexicalization. However, while these LLM-based methods excel at context, they are often stochastic and prone to identity fragmentation and merging. Our work extends this work by adding a graph-based solution to ensure relexicalization is mathematically bounded and deterministic yielding better temporal and non-temporal consistency.
\begin{table}[h]
\centering
\caption{Relex Baseline Inconsistency Examples}
\label{tab:relex-examples}
\begin{tabular}{|p{\dimexpr(\columnwidth-4\tabcolsep-3\arrayrulewidth)/2\relax}|p{\dimexpr(\columnwidth-4\tabcolsep-3\arrayrulewidth)/2\relax}|}
\hline
\textbf{Original Document} & \textbf{Relexicalized Document} \\
\hline
The patient \textbf{Alexander Miller}, born on \textbf{June 12, 1973, Tuesday}, reported worsening knee pain over the past six months. `It started around my \textbf{50th birthday},' they explained. \textbf{A Miller} \textbf{`Three years ago, at 47}, I had a minor cartilage tear, but physical therapy helped until now.' During the visit on \textbf{September 5, 2023}, the doctor \textbf{Jon Smith} noted decreased range of motion and recommended an MRI. `Given your family history of osteoarthritis in their late 50s,' the doctor \textbf{J Smith} added, `we should monitor this closely.'  & The patient \textbf{Michael Johnson}, born on \textbf{June 18, 1980, Tuesday}, reported worsening knee pain over the past six months. `It started around my \textbf{50th birthday last June},' they explained. \textbf{J Clarke} \textbf{`Three years ago, at early 50s}, I had a minor cartilage tear, but physical therapy helped until now.' During the visit on \textbf{September 5, 2023}, the doctor \textbf{Morgan Blake} noted decreased range of motion and recommended an MRI. `Given your family history of osteoarthritis in their late 50s,' the doctor \textbf{B Cooper} added, `we should monitor this closely.' \\
\hline
\end{tabular}
\end{table}

\section{System Overview}
\subsection{Motivating Example}
\label{sec:motivating_example}
Table \ref{tab:relex-examples} illustrates an example to demonstrate how inconsistencies in relexicalization could render the transformed documents useless for further analysis and insights. \\
\textbf{Temporal Inconsistencies:} The relexicalized document fails to maintain chronological consistencies relative to the patient's new birth date of June 18, 1980. In September 2023, the patient would be 43 years old, making references to a ``50th birthday'' and being in ``early 50s'' incorrect. \\
  
\textbf{Name Mapping Errors:} Doctor name abbreviations do not align between versions. Original uses ``Jon Smith'' and ``J Smith,'' which are different names of the same person - but relexicalized version mentions ``Morgan Blake'' and ``B Cooper'' who are two different identities. Similarly, the same patient (A Miller) is relexicalized to two different individuals ``J Clarke,'' and ``Michael Johnson''. \\

\textbf{Medical Implausibility:} Static phrases like ``family history of osteoarthritis in their late 50s'' remain unchanged despite the patient's age of 43, creating implausible clinical context. The visit date ``September 5, 2023'' also remains same but other temporal events are shifted.

\subsection{Clinical Knowledge Graph}
To solve the complex challenges of relexicalization, as illustrated in Section \ref{sec:motivating_example}, we introduce the concept of Clinical Knowledge Graph - a unified graphical representation of a patient's entire longitudinal history. \\

\begin{figure}[h]
    \centering
    \includegraphics[width=0.99\columnwidth, height=0.6\textheight, keepaspectratio]{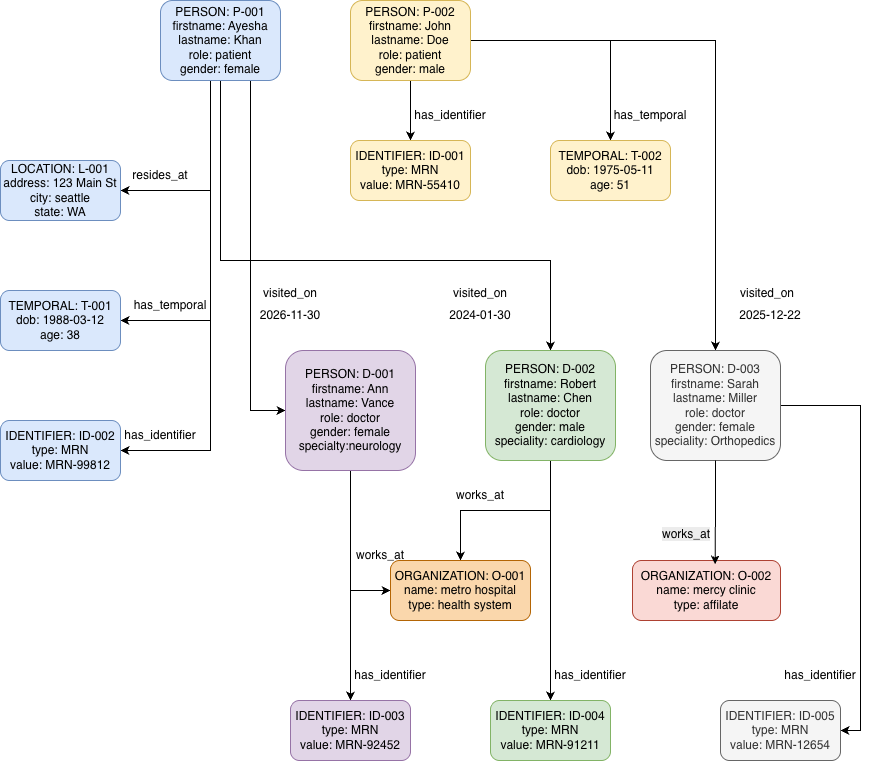}
    \caption{Typical \textbf{Clinical Knowledge Graph} structure}
    \label{fig:graph_example}
\end{figure}

\subsubsection{Knowledge Graph Structure}
The Clinical Knowledge Graph (CKG) is modeled as an attributed directed multigraph, defined by the structural tuple $G = (\mathcal{V}, \mathcal{E})$. Every node $v \in \mathcal{V}$ is assigned a specific categorical type $T(v)$ that belongs to one of six primary entity categories within an overarching taxonomy of clinical entity types $\mathcal{T}$. In addition to its entity type, each node maintains its own internal set of properties. The node types and properties are based on a list of 40 specific PII/PHI entities, as prescribed by HIPAA \cite{hhs_ocr_deid_guidance_2012}, \cite{Ahmedetal2020}. Additionally, a few temporal entities are also identified and used as node properties to represent temporal markers. Specifically, each node $v \in \mathcal{V}$ can be represented as a structured pair:
\[ v = \big( T(v), \mathcal{P}_v \big) \]
where $\mathcal{P}_v$ is a set of key-value pairs capturing the specific properties of that node:
\[ \mathcal{P}_v = \{ (k_1, w_1), (k_2, w_2), \dots, (k_n, w_n) \} \]
Here, $k_i$ denotes the property key and $w_i$ represents its corresponding data value.

\subsection{Methodology}
Our complete relexicalization pipeline (shown in Figure \ref{fig:Architecture}) consists of 3 stages as outlined below. \\
\textbf{Stage 1: Identification \& Classification of Entities}\par
\label{sec: PHI Identification}
\textbf{PHI/PII Identification}: The documents are first scanned to identify and tag HIPAA-specified PII/PHI entities and temporal entities (explicitly mentioned/inferred). The entity identification uses a combination of NER (named entity recognition) based methods \cite{oracle_oci_ner_2025} and LLM prompts. \\
\textbf{Input}: A raw text sequence $S \in \mathcal{\theta}$ \\
\textbf{Output}: Sequence of entities $E: $$E = \langle \text{text}, \text{beg}, \text{end}, \text{label} \rangle$ \\
\textbf{Entity Role Classification}: Next, to aid in disambiguation, each identified entity is assigned a role based on its context. This is critical for differentiating between, for example, a doctor and a patient who might share a similar name. The role association is done using an LLM prompt by providing it with the entity identified in Step 1 along with surrounding context. \\
\textbf{Stage 2: Initial CKG construction}:
The construction of the Clinical Knowledge Graph (CKG) begins by transforming unstructured clinical text into a dynamic, hybrid evidence network anchored to a central Patient ID (MRN,HEALTH\_ID,SSN,etc.). Following the initial Named Entity Recognition (NER) and Role Classification phases, each identified PII/PHI mentions (e.g., PERSON, ORGANIZATION) and temporal entity is instantiated as a graph node enriched with functional metadata/attributes (role, title, first\_name, street, etc.). 
The graph topology is constructed by combining probabilistic similarity edges derived from a tiered alignment over secondary attributes with deterministic relational edges extracted from primary identifier matches and explicit clinical or temporal dependencies. A typical structure of the CKG is shown in Figure \ref{fig:graph_example}.\\
\textbf{Stage3: CKG Integration and Multi-Stage Entity Resolution}: Continuing with the CKG construction process, incoming clinical data fields mapping specifically to \texttt{PERSON} entity categories---such as patients and clinicians---are integrated via a multi-stage resolution process. Other clinical entity types are handled independently by downstream context routers. To facilitate deterministic and probabilistic entity alignment, the complete taxonomy of recognizable clinical entity types $\mathcal{T}$ includes two disjoint subsets of identifier attributes: a primary identifier subset $\mathcal{I} \subset \mathcal{T}$ (e.g., SSN, MRN, HEALTH\_PLAN\_ID etc.) and a secondary identifier subset $\mathcal{A} \subset \mathcal{T}$ (e.g., first name, last name, gender, city, country, age etc.), such that $\mathcal{I} \cap \mathcal{A} = \emptyset$.


Given an incoming clinical entity record $u$, the framework first attempts an $O(1)$ deterministic match using only its primary identifiers. Let $\mathcal{I}_u \subseteq \mathcal{I}$ denote the set of active, observed primary identifiers extracted from $u$. If $\mathcal{I}_u \neq \emptyset$, the framework queries a globally indexed hash map of primary identifiers to find an existing target vertex $v \in \mathcal{V}$ via an exact-match index mapping function $\mathcal{M}: \mathcal{I} \to \mathcal{V}$:
\[
v^* = \mathcal{M}(k), \quad \text{where } k \in \mathcal{I}_u
\]
If a matching node $v^*$ is resolved via this primary index lookup, the entity is immediately aligned, bypassing downstream candidate generation and secondary scoring.

In cases where no deterministic match occurs (either because $\mathcal{I}_u = \emptyset$ or $\mathcal{M}(k)$ yields no result), the framework invokes a candidate generation step to generate a set of candidate vertices $\mathcal{C}_u \subset \mathcal{V}$. To accommodate typographical variances, abbreviations, and nickname initialisms (e.g., matching ``Sandeep Singh'' with ``Sandeep S.'') without scanning the global graph, this screening phase filters the graph using core demographic attributes viz.: first name ($\alpha$), last name ($\beta$), and gender ($\eta$). Formally, the candidate generation set $\mathcal{C}_u$ is defined as:
\[
\mathcal{C}_u = \left\{ v \in \mathcal{V} \;\middle|\; 
\begin{aligned}
  & f(\alpha_u, \alpha_v) \geq \tau_{\text{block}}, \\
  & f(\beta_u, \beta_v) \geq \tau_{\text{block}}, \\
  & \delta(\eta_u, \eta_v) = 1
\end{aligned}
\right\}
\]
where $\delta(x, y)$ represents a strict Kronecker delta function evaluating to 1 if the genders match exactly and 0 otherwise, $f$ denotes a string similarity metric (such as the Jaro-Winkler distance), and $\tau_{\text{block}}$ is a predefined screening threshold. This preliminary filtering phase prunes the search space via a lightweight similarity evaluation, restricting computationally intensive downstream calculations to a tightly constrained subset of candidate nodes.

Once the candidate pool $\mathcal{C}_u$ is isolated, the framework executes the secondary similarity matching function $\mathcal{R}(u, v) \in [0, 1]$ for each candidate node $v \in \mathcal{C}_u$. Let $\mathcal{A}_u, \mathcal{A}_v \subseteq \mathcal{A}$ denote the respective active secondary attributes extracted from the record and the localized candidate graph neighborhoods. We define $S_i \in [0, 1]$ as the local similarity score for any given secondary attribute $i \in \mathcal{A}$, evaluated using domain-appropriate distance metrics depending on the underlying data type. 

The global similarity scoring function over the candidate pool is modeled as a weighted aggregate over the intersecting subset of secondary identifiers:
\[
\mathcal{R}(u, v) = \frac{\displaystyle\sum_{i \in (\mathcal{A}_u \cap \mathcal{A}_v)} w_i \cdot S_i}{\displaystyle\sum_{i \in (\mathcal{A}_u \cap \mathcal{A}_v)} w_i}
\]
where $w_i \in \mathbb{R}^+$ denotes a predefined significance weight assigned to the $i$-th secondary identifier type. 

This structural approach provides a robust dual-layer defense. When ambiguous nodes emerge or lack direct primary identifiers, the localized execution of $\mathcal{R}(u, v)$ resolves them by examining both surface semantic meaning and structural position within the surrounding graph layout. This combined context allows the system to pull together matching variations into a single, unified profile while safely keeping distinct individuals with similar names separate based on their differing medical relationships. For longitudinal continuity across facilities, EHR normalization utilizes probabilistic record linkage and Master Patient Index (MPI) methodologies to map disparate local identifiers to a global patient ID. 

\begin{figure}[htbp]
    \centering
    \includegraphics[width=0.95\columnwidth, height=0.7\textheight, keepaspectratio]{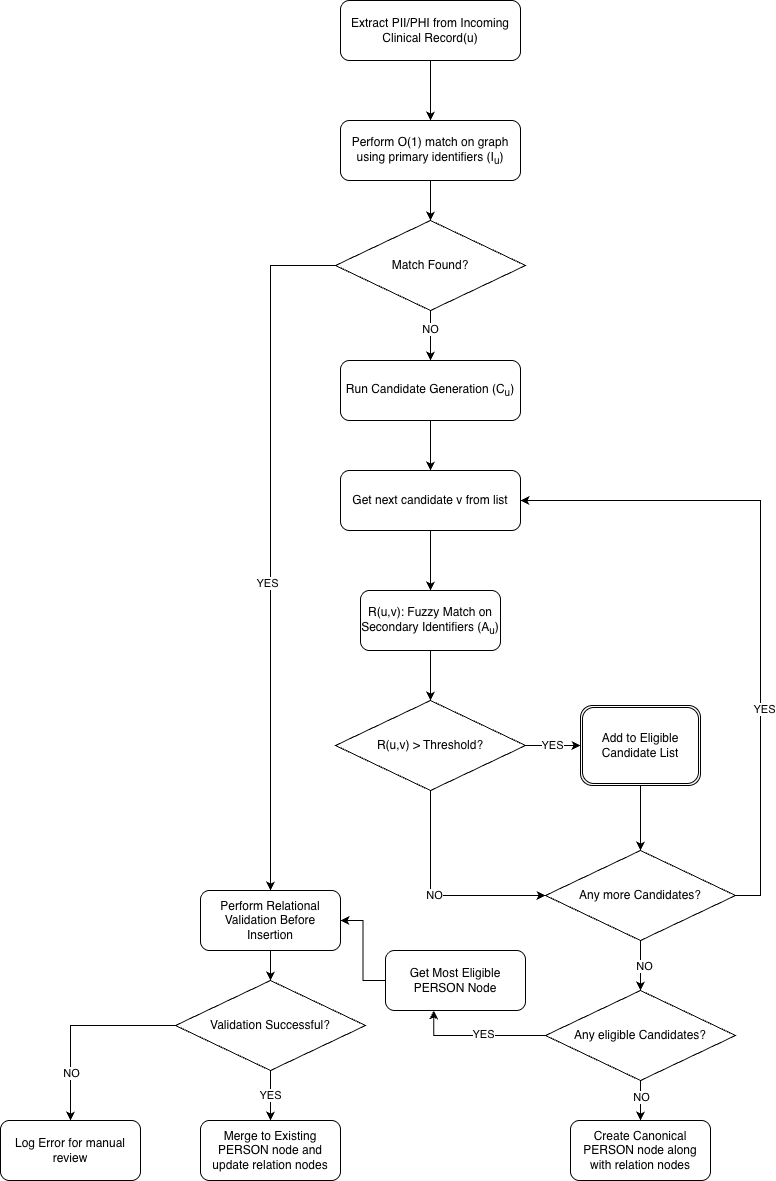}
    \caption{Entity Consistency Resolution Logic}
    \label{fig:entity_consistency_res}
\end{figure}

Once an edge resolution choice is settled, graph instantiation follows a split path based on the matching outcome. In the case of a match where a candidate yields a score $\mathcal{R}(u, v) \geq \gamma$ (threshold), the system selects the candidate with the highest similarity score, after which an additional relational validation is performed through ontological constraint checks. A successful validation triggers an automated update of the \texttt{PERSON} node properties and incremental synchronization of its relational neighbors based on the incoming clinical delta information. Validation failures indicate possible inaccuracies in the record and are logged for offline manual review. In the case of a fresh creation where $\mathcal{R}(u, v) < \gamma$ for all evaluated candidates in $\mathcal{C}_u$ (or if $\mathcal{C}_u = \emptyset$), a new \texttt{PERSON} node is instantiated, and its corresponding local relational subgraphs are structured natively from the properties in the source document. Figure \ref{fig:entity_consistency_res} details the decision logic used to query the graph and determine whether to link to an existing node or instantiate a new one.

\subsection{Temporal Relexicalization}
\label{sec: temporal_relex}
The primary objective of temporal relexicalization is to transform chronological entities within a clinical dataset into synthetic values while strictly preserving the relative temporal context. As identified in Section \ref{sec:challenges} (Temporal Inconsistency), Large Language Model (LLM) based relexicalization frequently introduces four distinct classes of logical inconsistencies. To mitigate these errors, we define the following expectations: \\
(i) \textbf{Chronological Consistency:} For events $e_i, e_j$ in patient's history, if $\text{date}(e_i) < \text{date}(e_j)$ in original, then $\text{date}'(e_i) < \text{date}'(e_j)$ after relexicalization \\
(ii) \textbf{Interval Consistency:} For date ranges $[T1, T2]$ (e.g., hospital admission to discharge), the relexicalized duration $[T1', T2']$ ~= $[T1, T2]$ after transformation \\
(iii) \textbf{Calendar Consistency:} The original and transformed date should have the same format (e.g. DD-MM-YY). Additionally, if original date mentions the day of the week, the relexicalized date must also reflect the correct day after transformation. \\
(iv) \textbf{Medical Plausibility:} For event $e$ with disease $d$, $\text{age}(e) \in [\text{age}_{\min}(d), \text{age}_{\max}(d)]$ after relexicalization. Additionally, seasonal constraints ($month(diag) \in seasonal\_window(diag)$) should hold.\\
\textbf{Temporal Identification and Repositioning} : Temporal entities (e.g., Age, DOB, Admission/Discharge dates) are identified and role-labeled in a single-stage LLM-prompting process, bypassing the multi-step approach in Sec \ref{sec: PHI Identification}. For relexicalization, these entities are shifted by a patient-specific offset ($\Delta$), generated via a hybrid engine (rule+LLM) to preserve temporal patterns while ensuring anonymity. 

\textbf{Rule-Based Constraint Layer:} This layer validates $\Delta$ for medical and chronological consistency using a global constraint table derived from: (i) ICD-10 guidelines; (ii) empirical patient cohort statistics; and (iii) expert clinical validation. 

\textbf{LLM Validation Layer:} A final LLM-assisted pass verifies document-level temporal correlations using the selected offset, ensuring longitudinal integrity. The LLM returns a confidence score $c \in [0, 1]$. Predictions with confidence below a pre-configured threshold fall back to alternative temporal shift values. \\

\begin{lstlisting}[
    basicstyle=\footnotesize\ttfamily,
    language=, 
    breaklines=true,
    %breakanywhitespace=true,
    % frame=tbLR ensures that if it splits across columns, 
    % top, bottom, Left, and Right lines are drawn for both fragments.
    frame=tbLR,
    frameround=ffff, % Keeps corners sharp and professional for IEEE
    xleftmargin=0pt,
    framexleftmargin=0pt,
    framesep=2pt,    % Adds clean internal padding between code and borders
    columns=fullflexible,
    breakindent=0pt,
    breakautoindent=false,
    postbreak={},
    rulecolor=\color{gray!60}, % Softens the border line color for a modern look
    captionpos=t,
    caption={LLM Prompt for generating Temporal Offset},
    label={lst:temp-offset}
]
You are an expert in temporal reasoning, clinical informatics, and medical data de-identification with deep knowledge of cross-document longitudinal consistency, medical plausibility verification, and adversarial privacy auditing.

Task: Given a patient longitudinal profile, a proposed patient-specific temporal offset(delta), and the relexicalized documents, verify that the transformation preserves (i) chronological coherence, (ii) longitudinal interval integrity, and (iii) clinical plausibility.

Assess the transformed timelines against the original records to identify any inconsistencies introduced by the temporal offset. A validation is successful only if all three criteria are satisfied across the longitudinal document set.

Return a strict JSON result containing: (a) an overall pass/fail status, (b) a confidence score, (c) the status and rationale for each validation criterion, and (d) any detected temporal or clinical contradictions. For each failure, provide a concise explanation of the affected event or constraint.


Constraints to Verify (ALL must pass for status=PASS):
1. Chronological Coherence (Cross-Document):
   - Ensure the linear ordering of events across ALL documents remains invariant after applying temporal offset.
   - Verify that sequential dependencies (e.g., Lab Test Ordered -> Lab Results Released -> Treatment Administered) do not contain inverted or overlapping timelines.

2. Longitudinal Interval Integrity:
   - Verify that the duration gaps between discrete documents or disparate clinical encounters match the original timeline perfectly (Zero-tolerance for multi-document drift).
   - Ensure age updates across multiple years of records advance realistically relative to the relexicalized Date of Birth (DOB).

3. Clinical Plausibility Preservation:
   - Audit the shifted timeline against pathophysiological boundaries. Reject if the offset places a chronic condition diagnosis (e.g., Pediatric Type 1 Diabetes vs. Adult Osteoarthritis) outside of its medically plausible age bracket.
   - Verify seasonal correlations: Seasonal diagnoses (e.g., winter influenza surges) must align with the post-shifted calendar months.

Confidence Score (c) Assessment Logic:
- c = 1.0: Flawless longitudinal alignment. All constraints perfectly met across the entire historical profile.
- 0.7 <= c < 1.0: Minor stylistic or formatting variance observed, but zero chronological or medical contradictions exist.
- c < 0.7: Any single logical failure, timeline inversion, or clinical implausibility automatically drops the score below the fallback threshold, necessitating an alternate offset recalibration (delta_p).

Prioritize rigorous, deterministic verification to prevent structural leakage or clinical incongruence in the final production output.
\end{lstlisting}
\textbf{Self-Improving Graph Robustness} : The CKG and longitudinal profile $G_p$ evolve incrementally by updating constraints, such as disease age-ranges ($\text{age}_{\min}$) to capture seasonality and comorbidity patterns, and triggering re-validation or offset recalibration ($\Delta_p$) whenever clinical contradictions arise to maintain global coherence.

\renewcommand{\algorithmiccomment}[1]{\hfill$\triangleright$ #1}
The temporal shifting and validation pipeline is executed systematically to ensure data consistency across patient records. The exact execution logic for managing these continuous and dynamic temporal calibrations is detailed in Algorithm~\ref{alg:temporal_relex}.

\begin{algorithm}[h]
\caption{Temporal Relexicalization}
\label{alg:temporal_relex}
\small
\begin{algorithmic}[1]
\Require Patient document $d_p$, patient ID $p$, CKG $G_p$, medical lookup table $G_{\text{med}}$, mapping database $M_p$
\Ensure De-identified document $d_p'$
\Statex
\Procedure{Temporal\_Relex}{$d_p, p, G_p, G_{\text{med}}, M_p$}
    \State $E \gets \text{ExtractTemporalEntities}(d_p)$ \Comment{Extract raw dates, ages, expressions}
    \Statex
    \If{$p \in M_p$}
        \State $\Delta_p \gets \text{RetrieveStoredOffset}(p, M_p)$
        \State $E_{\text{shifted}} \gets \text{ApplyShift}(E, \Delta_p)$
        \Statex
        \If{$\text{ValidateRuleLayer}(E_{\text{shifted}}, G_p, G_{\text{med}})$ \textbf{and} $\text{LLMScore}(E_{\text{shifted}}, G_{\text{med}}) > \text{threshold}$}
            \State \Comment{Case 1: Existing mapping maintains consistency; proceed with retrieved $\Delta_p$}
        \Else
            \State $\text{valid\_shifts} \gets \text{GenerateCandidates}([-\delta_T, +\delta_T])$ \Comment{Recalibrate}
            \State $\Delta_p \gets \text{FilterAndSelect}(\text{valid\_shifts}, G_p, G_{\text{med}})$
            \State $\text{StoreMapping}(p, \Delta_p, M_p)$ \Comment{Update stored offset mapping}
        \EndIf
    \Else
        \State $\text{valid\_shifts} \gets \text{GenerateCandidates}([-X, +X])$ \Comment{New patient - Case 2}
        \State $\Delta_p \gets \text{FilterAndSelect}(\text{valid\_shifts}, G_p, G_{\text{med}})$
        \State $\text{StoreMapping}(p, \Delta_p, M_p)$ \Comment{Store initial offset mapping}
    \EndIf
    \Statex
    \State $E_{\text{final\_shifted}} \gets \text{ApplyShift}(E, \Delta_p)$
    \State $d_p' \gets d_p$ \Comment{Initialize target document baseline}
    \Statex
    \ForAll{entity $e \in E_{\text{final\_shifted}}$}
        \State $\text{Validate}(e, \text{RuleLayer}, \text{LLMLayer})$
        \State $d_p' \gets \text{Replace}(d_p', e.\text{orig}, e)$ \Comment{Replaces original mention with shifted values}
    \EndFor
    \Statex
    \State $\text{UpdateGraphs}(p, E_{\text{final\_shifted}}, G_p, G_{\text{med}})$
    \State \Return $d_p'$
\EndProcedure
\end{algorithmic}
\end{algorithm}

\section{Experimental Setup}
The evaluation of G-RELIC utilized a heterogeneous corpus of clinical narratives and structured records, designed to reflect the multi-modal nature of modern Electronic Health Records (EHR). The total dataset encompasses 400 clinical documents spanning 50 unique patients and 15 clinicians across 5 healthcare organizations (covering 12 distinct clinical provider sites), exposing the model to highly diverse EHR formats, terminologies, and idiosyncratic dating conventions. The corpus includes longitudinal patient records spanning 3--18 documents, providing an extended temporal window for validating relexicalization algorithms across long clinical timelines.

\subsection{Corpus Composition and Stratification}
To evaluate the pipeline's versatility across different data models and linguistic complexities, the corpus is split into two primary structural formats: \textbf{Clinical Notes} and \textbf{Longitudinal Records}. Clinical Notes comprise unstructured or semi-structured JSON files containing summaries of clinician-patient encounters, including narrative prose following the SOAP (Subjective, Objective, Assessment, and Plan) framework. These documents present high lexical density and intricate temporal references specifically intended to test the pipeline's handling of complex narrative text. Conversely, Longitudinal Records consist of categorical JSON files providing a strict chronological view of patient histories across diverse clinical domains, such as \textit{Allergies}, \textit{MedicationRequests}, \textit{Immunizations}, and \textit{Conditions}. This dual-format architecture directly challenges the system's ability to maintain strict cross-document temporal consistency alongside dense prose analysis.

To capture varying degrees of real-world messiness and procedural edge cases, the dataset is stratified into three distinct tiers based on provenance, complexity, and manual curation. The \textbf{Gold tier} ($N=200$; 160 regular / 40 longitudinal) contains authentic production data sourced from the 12 external provider sites, featuring maximum real-world complexity, fragmented documentation, and highly heterogeneous date formats. The \textbf{Silver+ tier} ($N=50$; 40 regular / 10 longitudinal) consists of high-fidelity synthetic narratives explicitly authored by in-house clinicians to purposefully model rare and mathematically complex edge cases. Finally, the \textbf{Silver tier} ($N=150$; 100 regular / 50 longitudinal) comprises standard clinical documents manually curated by medical scribes and clinical researchers to represent routine, foundational clinical patterns. \footnote{Results are reported on the combined dataset due to small sample sizes for individual tiers.}

\subsection{Annotations}
Ground-truth PHI/PII annotations were performed and certified by an independent, third-party privacy organization compliant with HIPAA, PIPEDA, and GDPR frameworks. Temporal relexicalization accuracy was established using expert-validated mappings. \\
All experiments were deployed on a high-performance enterprise cloud infrastructure. The core architecture integrates the GPT-4.1 model with G-RELIC, a framework implemented as a labeled property graph using Neo4j. To ensure a deterministic and reproducible evaluation environment, the following LLM hyperparameters were used across all execution runs: temperature ($T = 0.1$), $\text{top\_p} = 0.9$, maximum tokens = 4096, frequency penalty = 0.0, and presence penalty = 0.0.

\section{Results}
Performance is evaluated across three dimensions—entity identity, temporal relationships, and privacy — using the following metrics:
\subsection{Evaluation Metrics}
\textbf{All Or Nothing Recall (AoN)}: We use a stricter version of the traditional recall metric viz. AoN recall to adhere to privacy benchmarks \cite{Scaiano2016AUF}. For any given entity type, a document receives a AoN score of 1 only if every instance is successfully relexicalized; otherwise, it receives a 0.\\
\textbf{Entity Consistency Score}: This metric quantifies the proportion of entity mentions that are consistently replaced with the same surrogate token throughout a document, ensuring predictability and reliability of the process: \begin{equation} \text{Entity Consistency (\%)} = \frac{ \text{Consistent Mentions}}{ \text{Total Mentions}} \times 100\% \end{equation}
\textbf{Temporal Consistency Score}: This score is calculated for the four types defined in Section \ref{sec: temporal_relex} as:
\begin{equation}
\begin{aligned}
\text{Temporal Consistency (\%)}\\
{}=\frac{\text{Consistent Instances}}{\text{Total Temporal Events}}\times 100\%
\end{aligned}
\end{equation}

\subsection{Performance and Efficiency}
\label{sec:perf_and_efficiency}
We compare the performance of G-RELIC against 3 baseline methods - viz. Rule-Based \cite{ijcai2019p689}, Random Shift \cite{Hripcsaketal2016} and Prompt Based (Rule+LLM) \cite{singh-etal-2025-redactor}. As demonstrated in Tables \ref{tab:entity_consistency} and \ref{tab:temp_consistency}, G-RELIC significantly outperforms all baseline systems across all consistency dimensions. In entity consistency, G-RELIC achieves an overall accuracy of 92.5\% ($1156$ consistent mentions out of $1250$), representing a 30.4 percentage point gain over the prompt-based baseline (62.1\%). The improvements are similar even if we consider individual entities. The performance gap is even more pronounced for temporal consistency; G-RELIC achieves 91.9\% overall, nearly doubling the performance of the LLM Prompt-based method (46.0\%). These results clearly indicate that for consistent relexicalization, a graph-based approach is superior to pure prompting. \\

Privacy benchmarks (Table \ref{tab:deid_performance}) confirm that G-RELIC successfully meets or exceeds AoN Recall targets for key entities ensuring that the relexicalized dataset is viable for downstream scientific research without compromising privacy. Furthermore, by replacing multi-pass strategies with a single-pass approach using GPT-4.1, G-RELIC achieves substantial efficiency gains: an \textbf{83\% reduction in input tokens, 84\% in output tokens, and 65\% improvement in P90 latency}.
\begin{table}[ht]
    \centering
    \footnotesize
    \resizebox{\columnwidth}{!}{%
    \begin{tabular}{l c c c c}
        \hline
        \textbf{Entity Class} & \textbf{Mentions} & \textbf{Rule} & \textbf{Prompt} & \textbf{G-RELIC} \\
         & \textbf{(N)} & \textbf{Based} & \textbf{Based} & \textbf{(Ours)} \\
        \hline
        PERSON          & 450   & 32.0 & 62.4 & 91.2 \\
        LOCATION        & 300   & 31.0 & 61.2 & 90.4 \\
        AGE             & 200   & 36.0 & 62.0 & 95.0 \\
        DOB             & 150   & 36.0 & 62.0 & 94.0 \\
        MARITAL\_STATUS & 150   & 35.0 & 62.7 & 92.0 \\
        \hline
        \textbf{Overall} & \textbf{1250} & \textbf{34.0} & \textbf{62.1} & \textbf{92.5} \\
        \hline
    \end{tabular}
    }
    \caption{Entity Consistency Score (\%) Performance.}
    \label{tab:entity_consistency}

    \vspace{1.5em} 

    \resizebox{\columnwidth}{!}{%
    \begin{tabular}{lllll}
        \hline
        \textbf{Consistency}  & \textbf{Mentions} & \textbf{Random} & \textbf{Prompt} & \textbf{G-RELIC} \\
        \textbf{Metric} & \textbf{(N)} & \textbf{Shift} & \textbf{Based}  &\textbf{(Ours)} \\
        \hline
        Chronological          & 220  & 26.0 & 40.0  & 81.8 \\
        Calendar Format        & 200  & 89.1 & 60.0  & 98.2 \\
        Interval               & 180  & 11.8 & 20.9  & 97.8 \\
        Clinical Plausibility      & 185  & 31.9 & 63.2  & 88.1 \\
        \hline
        \textbf{Overall} & \textbf{785} & \textbf{39.7} & \textbf{46.0} & \textbf{91.9} \\
        \hline
    \end{tabular}
    }
    \caption{Temporal Consistency Score (\%) Performance.}
    \label{tab:temp_consistency}

\end{table}

\begin{table}[ht]
\centering
\resizebox{\columnwidth}{!}{%
\begin{tabular}{lcc}
 \hline
\textbf{Entity Class} & \textbf{Prompt-Based} & \textbf{G-RELIC} \\
\textbf{(Target AoN)} & \textbf{Precision/Recall (\%)}   & \textbf{Precision/Recall (\%)} \\ 
 \hline
\midrule
\midrule
\multicolumn{3}{l}{\textit{\textbf{Proprietary Dataset}}} \\
PERSON ($\ge$ 97.5\%)    & 91.00 / 96.00 & 92.00 / 97.00 \\
LOCATION ($\ge$ 80.0\%)  & 77.00 / 100.00 & 75.00 / 100.00 \\
AGE ($\ge$ 80.0\%)      & 74.00 / 90.00 & 74.00 / 92.00 \\
ID ($\ge$ 80\%)        & 85.00 / 96.00 & 87.4 / 97.00 \\
DATE ($\ge$ 80\%)      & 89.40 / 96.00 & 89.2 / 97.00 \\
 \hline
\textbf{Overall}         & \textbf{83.28 / 89.00} & \textbf{83.52 / 92.00} \\
 \hline
\multicolumn{3}{l}{\textit{\textbf{Public Dataset (N2C2 2014\cite{stubbs2015automated})}}} \\
PERSON                   & 99.12 / 95.95 & 99.00 / 96.93 \\
LOCATION                 & 100.00 / 78.55 & 100.00 / 82.74 \\
AGE                      & 99.30 / 89.87 & 99.00 / 91.64 \\
ID                       & 66.67 / 92.75 & 74.81 / 96.21 \\
DATE                     & 99.88 / 94.95 & 99.00 / 95.00 \\
 \hline
\textbf{Overall}             & \textit{\textbf{97.90 / 91.59}} & \textit{\textbf{98.10 / 92.15}} \\ 
 \hline
\bottomrule
\end{tabular}
}
\caption{Unified De-identification and PHI/PII Detection Performance Across Proprietary and Public Datasets}
\label{tab:deid_performance}
\end{table}

\subsection{PHI/PII performance}
While G-RELIC prioritizes consistency, Table \ref{tab:deid_performance} confirms it does not regress on critical privacy benchmarks. The framework keeps the de-identification precision largely intact while achieving a slight improvement in AoN recall. It is important to note that this observed difference in recall should not be attributed to the proposed graphical consistency method itself, but rather to a more rigorously optimized prompt tuning strategy deployed for the underlying de-identification layer. Overall, G-RELIC successfully meets or exceeds AoN Recall targets for most key entities, ensuring that the relexicalized dataset remains highly viable for downstream scientific research without compromising patient privacy. Additionally, by replacing the multi-chunk, multi-pass strategy of \cite{singh-etal-2025-redactor} with a single-pass approach and leveraging GPT-4.1's superior language comprehension skills and larger context window, G-RELIC achieved substantial efficiency gains, namely an \textbf{83\% reduction in input token usage, an 84\% reduction in output token usage, and a 65\% improvement in P90 latency}.

\subsection{Limitations}
While the experimental results demonstrate robust performance and significant efficiency gains, a granular error analysis reveals specific scenarios or edge cases where entity resolution can be further refined, highlighting opportunities for future improvement. \\
The system is susceptible to fragmentation when primary identifiers are missing or inaccurately recorded across disparate clinical databases. In these instances, if the secondary identifiers also fail to meet the required similarity threshold $\gamma$, the framework will not be able to establish a deterministic link. This may result in the system instantiating a duplicate canonical node instead of a merge, resulting in degraded longitudinal consistency after relexicalization. \\
Isolated temporal discrepancies may also arise when age or demographic data is captured inconsistently within clinical narratives. If secondary attributes are insufficient to confirm patient identity with high confidence, the system cannot reliably anchor the record to an existing profile, resulting in creation of redundant nodes. \\
Our future work will focus on optimizing linkage constraints for these low-confidence, high-uncertainty scenarios to improve consolidation accuracy.

\subsection{Operational Metrics and Efficiency} 
Following offline verification, G-RELIC was deployed to a live production environment to evaluate its scalability and stability under real-world workloads. Because security and compliance guidelines strictly prohibit direct human review or "eyes-on" access to live patient charts, operational metrics were collected exclusively via automated, privacy-preserving telemetry dashboards. Over a 7-day deployment window, the system processed a total of $1,176,642$ relexicalization requests with a $98.7\%$ success rate (1,161,542 successful vs. 15,100 failed). To support this workload, the system executed 4,754,301 LLM inferences, processing a cumulative volume of 48.4 million input tokens and 997,000 generated output tokens. Despite this high throughput, the system maintained a highly optimized cost profile, totaling $\$105$ for the period. Latency metrics remained within acceptable bounds for real-time applications: the LLM response time showed a median ($P_{50}$) of $656$ ms, and a $P_{99}$ of $1.63$ s, while the end-to-end system response time yielded a $P_{50}$ of $5.10$ s and a $P_{99}$ of $7.36$ s.

\section{Conclusion and Future Work}
G-RELIC is a novel clinical relexicalization framework providing mathematically bounded, deterministic consistency for high-fidelity datasets. Building upon this framework, we envision several paths for extending its capabilities. A primary direction involves the integration of Graph Neural Networks (GNNs) to move beyond rule-based constraints, allowing the system to learn and generalize disease-specific temporal properties directly from large-scale medical corpora. We also aim to enhance the framework's privacy profile by integrating Differential Privacy guarantees on top of our deterministic shifts, providing a hybrid model of utility and formal privacy. \par
In terms of algorithmic refinement, we will continue to evolve our entity resolution logic, specifically by tuning linkage constraints to better handle low-confidence matches and thereby enhance overall consolidation accuracy. Furthermore, we plan to explore multilingual relexicalization to support non-English EHR systems beyond North America. Finally, a large-scale evaluation in production healthcare environments will be conducted to assess the framework’s real-world deployment challenges and its measurable impact on clinical decision-support research.

\section{Acknowledgements}
We are deeply grateful to our colleagues in Oracle Health AI for their ongoing collaboration and insightful perspectives. Special thanks to Salil Rajeev Joshi, Kiran Rama, Bhagya Hettige, Praphul Singh, Neil Hauge, Brad Jacobs, Mark Johnson, Krishnaram Kenthapadi, Shirley Liu, Laurent Boue, Amitabh Saikia, Vishal Vishnoi, and Raefer Gabriel for their invaluable input, guidance and support throughout the course of this research. Furthermore, we thank the anonymous reviewers for their constructive comments and thoughtful suggestions, which significantly enhanced the final quality of this manuscript.

\bibliographystyle{IEEEtran}
\bibliography{grelic}

\end{document}